\documentclass[sigconf]{acmart}

\usepackage[utf8]{inputenc} 
\usepackage[T1]{fontenc}    
\usepackage{url}            
\usepackage{booktabs}       
\usepackage{amsfonts}       
\usepackage{nicefrac}       
\usepackage{microtype}      
\usepackage{mathtools}

\usepackage[toc,page]{appendix}
\usepackage{url}
\usepackage[neverdecrease]{paralist}
\usepackage{graphicx}
\usepackage{booktabs}
\usepackage{graphics}
\usepackage{tabularx}
\usepackage{xspace}
\usepackage{graphicx}
\usepackage{booktabs}
\usepackage{mdwlist}
\usepackage{multirow}
\usepackage{amsmath}
\usepackage{mathtools}
\usepackage{enumitem}
\usepackage{calc}
\usepackage[neverdecrease]{paralist}
\usepackage[ruled,noend]{algorithm2e}
\usepackage{booktabs}
\usepackage{diagbox}
\usepackage{colortbl}
\usepackage{subcaption}
\usepackage{makecell}
\usepackage{verbatim}
\usepackage{amsmath}
\usepackage{color,soul}

\usepackage[figureposition=bottom,tableposition=top]{caption}

\usepackage{amssymb}%
\usepackage{pifont}%
\newcommand{\urlwofont}[1]{ \urlstyle{same}\url{#1} }

\renewcommand{\vec}[1]{\mathbf{#1}}
\newcommand{\para}[1]{\paragraph{\textnormal{\textbf{#1}}}}

\usepackage[utf8]{inputenc} 
\usepackage[T1]{fontenc}    
\usepackage{url}            
\usepackage{booktabs}       
\usepackage{amsfonts}       
\usepackage{nicefrac}       
\usepackage{microtype}      
\usepackage{soul}

\title{\emph{ALEX}: \emph{A}ctive \emph{L}earning based Enhancement of a Model's \emph{EX}plainability
}

\begin{document}


\author{Ishani Mondal}
\affiliation{%
 \institution{Indian Institute of Technology,\\
 Kharagpur, India}
}
\email{ishani.mondal@iitkgp.ac.in}

\author{Debasis Ganguly}
\affiliation{%
 \institution{IBM Research Europe\\Dublin, Ireland}
}
\email{debasis.ganguly1@ie.ibm.com}

\begin{abstract}
An active learning (AL) algorithm seeks to construct an effective classifier with a minimal number of labeled examples in a bootstrapping manner. While standard AL heuristics, such as selecting those points for annotation for which a classification model yields least confident predictions, there has been no empirical investigation to see if these heuristics lead to models that are more interpretable to humans. In the era of data-driven learning, this is an important research direction to pursue. This paper describes our work-in-progress towards developing an AL selection function that in addition to model effectiveness also seeks to improve on the interpretability of a model during the bootstrapping steps. Concretely speaking, our proposed selection function trains an `explainer' model in addition to the classifier model, and favours those instances where a different part of the data is used, on an average, to explain the predicted class. Initial experiments exhibited encouraging trends in showing that such a heuristic can lead to developing more effective and more explainable end-to-end data-driven classifiers. 
\end{abstract}


\maketitle

\section{Introduction}

A supervised learning approach (e.g. classification)
estimates a functional dependence
of the form $\theta: X \mapsto Y$
between a set, $X=\{\vec{x}\}, \vec{x} \in \mathbb{R}^d$, of data instances (strictly speaking, a $d$-dimensional encoding of data instances), and a set, $Y=\{y\}, y \in \mathbb{Z}_c$ of $c$ categorical value labels. 
This functional dependence is estimated from \emph{training examples}, i.e. samples of pairs from the set $X \times Y$. With a set comprising $M$ such pairs, $\{(\vec{x}_i, y_i)\}_{i=1}^M$, a parameterized classification model then estimates this function as a map from the training pairs to
a $c$-class probability distribution function,
$\hat{\theta}: \{(\vec{x}_i, y_i)\}_{i=1}^M \mapsto \Delta_{c-1}$.
%
Obviously, the error in this parametric approximation, $\hat{\theta}$ (of the true functional dependence, $\theta$), is expected to be smaller with increasing number of pairs in the training set, $M$, i.e.,
$\hat{\theta} \to \theta$ as $M\to \infty$.
However, since it is difficult to obtain a large number of training examples due to the annotation cost involved in generating such labeled data pairs, a standard solution is to
employ a methodology, called \emph{active learning} \cite{settles2009active}, 
to select a small subset of instances that are estimated to be good candidates to be labeled (annotated) among a very large number (pool) of unlabeled instances.

An \textbf{active learning} (AL) algorithm works in a bootstrapping manner, i.e., it starts with a small \emph{seed set}, $S$ of $M_0$ labeled data instances, $S = \{(\vec{x}_i, y_i)\}_{i=1}^{M_0}$ from a universal set of instances, $U$, i.e., $S \subset U$. It then employs a selector function of the form $\sigma: \vec{x} \mapsto \{0,1\}$ ($\vec{x} \in U-S$), which decides if an yet unlabeled instance is to included in the next batch of the annotation process, after which the seed set, $S_0$ is expanded to $S_1=\{(\vec{x}_i, y_i)\}_{i=1}^{M_1}$ and so on for $p$ successive steps.
Different selection functions, as parts of different AL strategies, thus lead to different classifier models. This is because the parameterized classifier model $\hat{\theta}$
depends on the selector function and the number of AL steps ($p$), i.e.,
$\hat{\theta}(\sigma, p): U
\xrightarrow{p}
S_p \mapsto \Delta_{c-1}$.
%
The selection function of an AL method (say $A$), $\sigma_A$ is more effective than another (say $B$) if
$|\theta - \hat{\theta}(\sigma_A,p)| < |\theta - \hat{\theta}(\sigma_B,p)|$, or in other words, after an identical number of steps (and labeled instances), method $A$ leads to better classification effectiveness than method $B$.

Existing approaches to selection functions of AL methods employ a wide range of different heuristics, such as that of
\emph{uncertainty sampling} \cite{uncertain_sampling}, which involves selecting the unlabeled instances for which the classifier trained on the labeled dataset at the $j^{th}$ bootstrapping step, $S_j=\{(\vec{x}_i, y_i)\}_{i=1}^{M_j}$,
yields the least certain predictions. While such heuristics have been empirically shown to work well towards developing effective classification models with a small number of data instances (chosen by the heuristic at each step of bootstrapping), there has been no empirical investigation (to the best of our knowledge) to see how \emph{interpretable} these models are.
%

In a data-driven model, due to the absence of explicitly annotated features, not all candidates (determined by the choice of the selection function) are expected to lead to interpretable models. For example, consider the task of developing a classifier for identifying hand-written digits, e.g. the MNIST \cite{lecun-mnisthandwrittendigit-2010}. For feature-driven statistical approaches, features computed from a character image, such as the slant or the continuity in the strokes, could be useful to interpret the predictions of the final model (constructed with AL based selection of labeled data instances). However, for a data-driven approach, it may be rather difficult to interpret the predictions in terms of these human perceivable features.

Existing per-instance model explanation approaches (e.g., LIME \cite{Ribeiro} or Shap \cite{SHAP}) interpret data-driven models in terms of soft-attention weights assigned to the data components of each instance.
The overall explanation for a model constitutes selected components from a subset of data instances with high values of these soft attention weights (e.g., regions within images or terms within documents).
Since the overall model explanation largely depends on the labeled data instances themselves, it is likely that some AL strategies may eventually lead to more \emph{explainable} models than others (in the sense that the soft attention weights for some sets of instances would better correlate with human perception).  

This paper reports our work towards exploring a novel AL strategy for enhancing the interpretability of a bootstrapped model in addition to the standard AL goal of improving its effectiveness. In particular, we propose a selection function strategy that seeks to make the final data-driven bootstrapped model more interpretable in terms of the estimated weights (\emph{explanation vector}) of the data components. Experiments indicate that in contrast to standard AL selection functions, our proposed selection methodology yields more \emph{effective} and \emph{explainable} models. 

\section{Related Work}

Selection heuristics for active learning (AL) can broadly be categorized into two approaches. The first among these is the \textbf{input distribution} based method which rely only on the similarities of the data instances annotated so far with the ones that are not yet annotated \cite{fujii-etal-1998-selective,preclustering}. The second category corresponds to \textbf{model uncertainly} based approaches, where instances selected for annotation are those for which the current model makes prediction with least confidence (e.g. they being the closest to the current decision boundary) \cite{uncertain_sampling,seqlabeling}.
The selection corresponding to the first class of methods is usually faster, involving pairwise distance computation, e.g., \cite{fujii-etal-1998-selective} selects points that are least similar to the labeled instances (seeking to maximize diversity), and \cite{preclustering} selects a set of core points from each cluster avoiding the outliers aiming hence to minimize uncertainty (of cluster membership). 
The second class of AL methods is usually more effective because these methods additionally leverage information from the class posteriors. A close to uniform posterior probability distribution over $c$ classes, $\delta_{c-1}$, represents a situation when a model is uncertain about the predicted class. By greedily selecting the set of instances that yield most uncertain posteriors, the uncertainty sampling (US) algorithm \cite{uncertain_sampling} aims to maximize the robustness of predictions. The US algorithm has been shown to be effective for sequence labeling as well \cite{seqlabeling}.   

Among research that relate to both model interpretability and AL include those of \cite{ghai2020explainable}, which
conducts an user study to find if model explanations affect the user annotations, and \cite{Ial}, where the aim was to apply a model explanation method to help the annotators construct a dataset that is able to make more fair (less biased) predictions. Our work is different in the sense that neither do we conduct a real user study \cite{ghai2020explainable}, nor do we aim to construct a more balanced dataset of labeled examples \cite{Ial}. Instead, our proposed selection function uses instance-wise explanations to combine both uncertainty of posteriors and model interpretability seeking to construct models that are both effective and interpretable using a minimal number of data annotations.

\section{Explanation-based Selection}

\para{Transforming Data to Explanations}
The first step of our proposed selection function is to \emph{transform} each input instance $\vec{x} \in \mathbb{R}^d$ to a different space representing its feature importance or explanation vector (also known as soft attention weights in the literature). Generally speaking, given a particular data instance $\vec{x}$ and a parameterized (trained) model $\theta: \vec{x} \mapsto y$, an instance-wise explanation method (e.g. LIME \cite{Ribeiro}, L2X \cite{L2X} or Shap \cite{SHAP}) first samples other similar data instances, $\vec{z} \in N(\vec{x})$, from its neighborhood and then learns a \emph{local view} of the global model by leveraging the predictions of $\theta$ on these \emph{local} instances.
The instance-wise explanation objective is to make the local model $\theta_{\vec{x}}$ \emph{closely approximate} the global model $\theta$. The objective is to minimize a loss of the form
\begin{equation}
\mathcal{L}(\vec{x},\theta; \phi) = \!\!\!\!\!\!\sum_{\vec{z} \in N(\vec{x})}
\!\!\!(\theta(\vec{z}) - \phi\cdot\vec{z})^2,\,\,
N(\vec{x})= \{\vec{x}\odot\vec{u}: \vec{u}\sim \{0,1\}^d\} \label{eq:instance_exp},
\end{equation}
where $\phi\cdot\vec{z}$ is a parameterized linear representation of the local function $\theta_{\vec{x}}$. The neighborhood function is approximated by selecting arbitrary subsets of features of the current instance $\vec{x}$ (of the form $\vec{x}\odot \vec{u}$, where $\vec{u}$ is a random bit vector of size $d$). A set of data instances with diverse explanation weights (e.g., as computed with an approximate set cover in \cite{Ribeiro}) may then constitute a global explanation unit for a model.

In particular for our experiments, we make use of the Shap approach for computing explanation weights, which while applying Equation \ref{eq:instance_exp}, specifically computes the importance of each feature with the help of Shapley values \cite{shapley}. The Shapley value of a feature considers its inclusion/exclusion effect on the model predictions by computing the relative differences in the posteriors.

In the context of our work on AL, given a current model $\theta_j$, trained with $S_j$ (the labeled dataset at the $j^{th}$ bootstrap iteration), Equation \ref{eq:instance_exp} is first applied on each instance $\vec{x} \in S_j$ to learn the parameters $\phi$. Next, we obtain an explanation vector $e(\vec{x}) = \phi.\vec{x}$
for $\vec{x}$ either in the labeled or in the unlabeled set of instances. We now describe how we use the explanation vector, $e(\vec{x})$ of an instance $\vec{x}$ in our selection methodology. 



\para{Model Interpretation Diversity}

During each step of AL bootstrapping, the objective is to select the next batch of instances for annotation. We hypothesize that the data instances which are the most difficult ones to explain with the help of the current model (trained on the available labeled data set) are the ones that should be annotated in order to maximize the interpretability of the model in the subsequent iterations.
Specifically, we measure the difficulty of explaining an unannotated instance, $\vec{x_u} \in U$, with the explanation model trained on the currently available labeled data, $S$ (Equation \ref{eq:instance_exp}), as the average KL-divergence between $\vec{x_u}$ and the explanation vectors estimated for instances in $S$.
We then select the points from $U$ with the highest average divergence values.

A high value of the average divergence in explanation indicates that there may a substantial disagreement on which parts of the data instances (e.g. regions in images) are likely to be useful to explain the model output between the two different sets of (currently) labeled ($S$) and unlabeled ($U$) instances.
The purpose of selecting the instances with the highest divergence values gives the annotation process a chance to label the points that potentially led to this difference in model interpretability.
Additionally, to take into account model uncertainty,
instead of computing the average KL-divergence for each point in $U$, we restrict the computation to a candidate set only, comprising the points with maximum uncertainty in the class posteriors.
Algorithm \ref{algo:exal} outlines
our proposed approach.

It is worth mentioning that the local instance-wise explanation vectors during each AL bootstrapping step contribute to a global set of diverse and potentially more interpretable explanation units (this is different from the approximate set cover based construction of explanation units proposed in \cite{Ribeiro}).

\begin{algorithm}[t]
\small
\DontPrintSemicolon
\KwIn{A set of unlabeled data instances, $U=\{\vec{x}\}$.}
\KwIn{A seed set of $M_0$ labeled data instances $S_0=\{(\vec{x},y)\}_{i=1}^{M_0}$.}
\KwIn{$b$: \#instances to select for annotation at each step.}
\KwIn{$k$: Size of the set of candidate data instances to consider for selection based on uncertainty sampling ($k > b$).}
\KwIn{$p$: \#iterations of active learning bootstrapping.}
\KwOut{An expanded set of $S$ of labeled data instances.}
\Begin{
$j \gets 1$, $S\gets S_0$\;
\While {$j \leq p$} {
    Train $\theta: \{\vec{x}_i\}_{i=1}^{|S|}\mapsto \{y_i\}_{i=1}^{|S|}$\;
    Train explanation model $\phi$ on $\vec{x} \in S$ and $\theta$ with Equation \ref{eq:instance_exp}\;
    \ForEach {$\vec{x} \in U$} {
        $s(\vec{x}) = \max_{l=1}^{c}\sigma(\theta\cdot\vec{x})_l$
        \tcp{softmax to compute the most likely posterior class (\#classes=$c$).}
    }
    $C \gets$ Set of $k$ instances from $U$ with the lowest $s(\vec{x})$ scores\;
    \tcp{Compute mean KL-Div between the explanation vector of an unlabeled instance and the labeled instances.}    
    \ForEach {$\vec{x_u} \in C$} {
        $\bar{d}(\vec{x_u}) \gets 0$\;
        \ForEach {$\vec{x} \in S$} {
            $d(\vec{x_u},\vec{x})=\mathrm{KLD}(\phi\cdot\vec{x_u},\phi\cdot\vec{x})$
            $\bar{d}(\vec{x_u})\gets \bar{d}(\vec{x_u}) + d(\vec{x_u},\vec{x})$            
        }
        $\bar{d}(\vec{x_u}) \gets \bar{d}(\vec{x_u})/|S|$
    }    
    $\Delta S \gets$ $b$ instances from $C$ with the highest $\bar{d}(\vec{x_u})$ values\;  
    $S\gets S \cup \Delta S$, $U \gets U - \Delta S$\;
}
}
\caption{ALEX: AL to enhance EXplainability  \label{algo:exal}}
\end{algorithm}

\section{Evaluation}

\subsection{Experiment Setup}

\para{Datasets}
As the classification task, we experiment with two standard image classification datasets, namely MNIST \cite{lecun-mnisthandwrittendigit-2010} and Fashion-MNIST (FMNIST) \cite{fmnist}. Both these datasets comprise grayscale images corresponding to one of $10$ classes. While the images in the MNIST dataset corresponds to those of hand-written digits ($\{0,\ldots,9\}$), the FMNIST ones correspond to those of clothing or footwear, e.g. t-shirts, sneaker etc. The number of images in the training and the test splits of the two datasets are identical, i.e., $60K$ train and $10K$ test images.

\para{Baselines}
The first among the baseline AL approaches is a relatively simple yet a standard baseline used in the AL literature, is the \emph{random sampling} (\textbf{RS}) based selection, which involves selecting instances from $U$ at random to be included for annotation in the next batch \cite{settles2009active}.  
Our next baseline is the popular AL approach of \emph{uncertainty sampling} (\textbf{US-P}) \cite{uncertain_sampling,seqlabeling}, which selects those points for which the classifier constructed with the current version of the incrementally growing labeled dataset yields the least confident predictions. We also use a discriminating version of the uncertainty sampling baseline, which instead of the posteriors makes use of the distances of the points from the classifier boundaries (margins) (\textbf{US-M}) \cite{margin-US}.  
We also employ the input distribution based approach of \emph{density weighted} (\textbf{DW}) estimation, which involves selecting those points for annotations which are least similar to the currently labeled instances. In particular, we follow the method proposed in \cite{fujii-etal-1998-selective}, which first clusters the set of input data instances into $c$ clusters (where $c$ is the number of classes for the classification task, which in our case is 10) to compute the distance of an unlabeled point from the centroids of the $c$ clusters.

\para{Experiment Workflow}

We use the training split of the datasets ($60K$ images) to incrementally construct the labeled pool of images. Each AL method starts with a seed set of labeled images randomly chosen from the training split of the datasets ($q$ number of images from each class, $q$ being a parameter which we vary in our experiments). With $q$ number of examples in each class (out of $c$), the initial seed set size, $|S|=qc$, i.e. $10q$, in our case.
For a fair comparison, we use an identical initial seed for each AL method, and execute $p=10$ steps for each AL method.

During each step of an AL method, we select $b=|S|$ points (see Algorithm \ref{algo:exal}) to be added to $S$, i.e. after $p$ AL steps, the total number of labeled points become $|S_p|=p|S_0|$. Specific to our proposed method, ALEX, the size of the candidate set was set to 3 times that of the batch size in our experiments, i.e. $k=3b$.  
For the final evaluation of an AL bootstrapped model,
we use the test split of the datasets. As the effectiveness metric, we compute the accuracy values obtained with models trained with each step ($p=1\ldots,10$) of AL. For model interpretability, we present a visual analysis of the explanation vector weights of the final model ($p=10$) for a sample set of images from the two datasets.

\para{Implementation Details}
A Keras implementation of our image classification model constitutes a standard single-layer 2D convolution ($3 \times 3$ kernel with dimensions 32), a fully connected layer ($128$ dimensional) with RELU activation \cite{ReLU}, and a softmax ($c=10$ dimensions) activation layer at the output.
During each step, the classifier model for each AL method was trained with
the Adam optimizer employing the cross-entropy loss \cite{cross-entropy}. 

\begin{table}[t]
\centering
\small
\caption{Accuracy on the test-set after $p=10$ bootstrapping steps obtained with a number of AL approaches with different seed sizes ($q$ instances for each class to initialize the seed).
\label{tab:results}
}
\begin{tabular}{l ccc ccc}
\toprule
AL & \multicolumn{3}{c}{MNIST} & \multicolumn{3}{c}{FMNIST}\\
\cmidrule(r){2-4}
\cmidrule(r){5-7}
Method & $q=1$ & $q=5$ & $q=10$ & $q=1$ & $q=5$ & $q=10$\\
\midrule
RS &
0.5294 & 0.7753 & 0.8368 &
0.5089 & 0.6051 & 0.7476 \\
US-P &
0.5355 & 0.8136 & 0.8533 &
0.6327 & 0.6902 & 0.7724 \\
US-M &
0.5776 & 0.8085 & 0.8519 & 
0.4970 & \textbf{0.7240} & 0.7662 \\
DW &
0.5848  & \textbf{0.8362} & 0.8733 & 
0.5190 & 0.7179 & 0.7772 \\
ALEX &
\textbf{0.6064} & \textbf{0.8362} & \textbf{0.8811} &
\textbf{0.6491} & 0.6845 & \textbf{0.7887} \\
\bottomrule
\end{tabular}
\end{table}

\subsection{Results with Concluding Remarks}

\begin{figure}[t]
    \centering
\begin{subfigure}[t]{.32\columnwidth}
    \centering
    \includegraphics[width=\columnwidth]{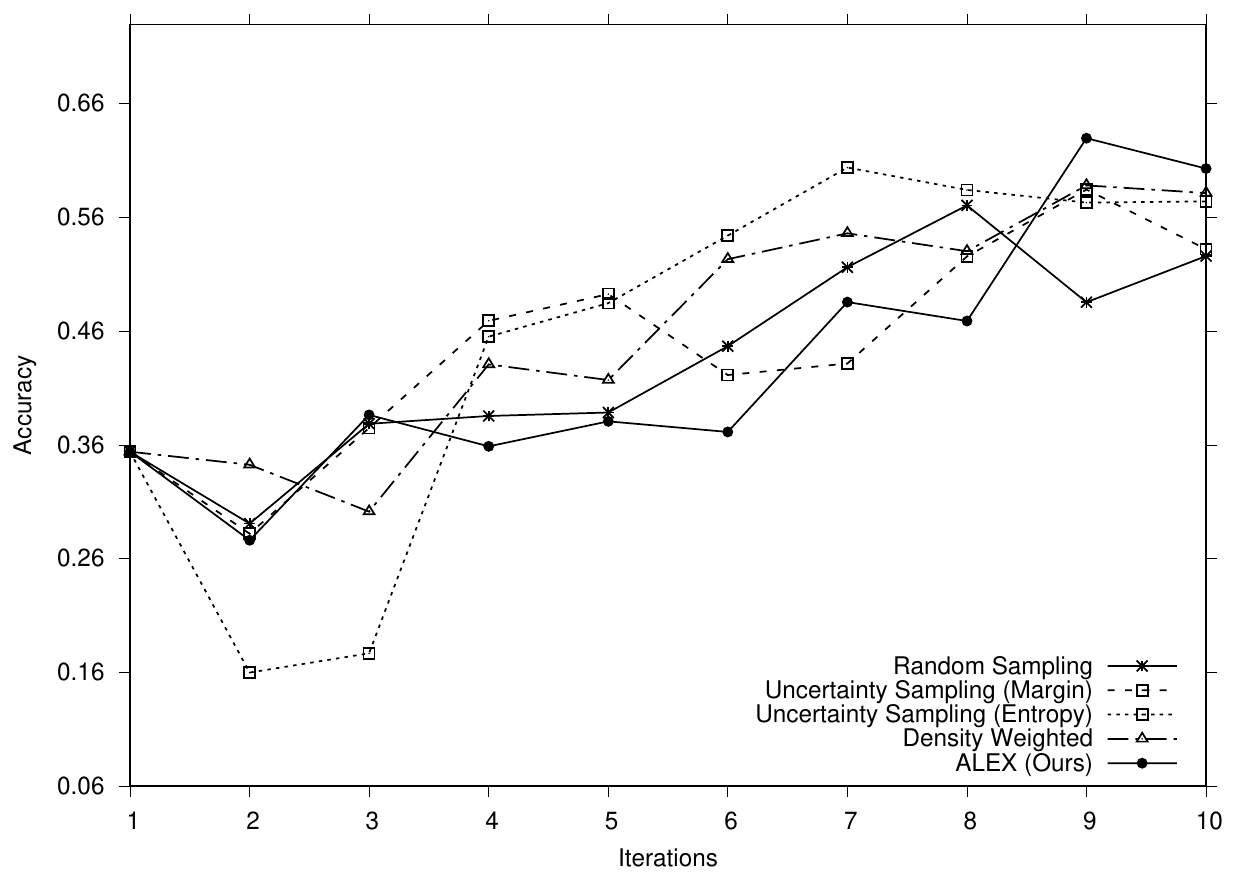}
    \caption{$q=1$ ($|S|=10$)}
\end{subfigure}
\begin{subfigure}[t]{.32\columnwidth}
    \centering
    \includegraphics[width=\columnwidth]{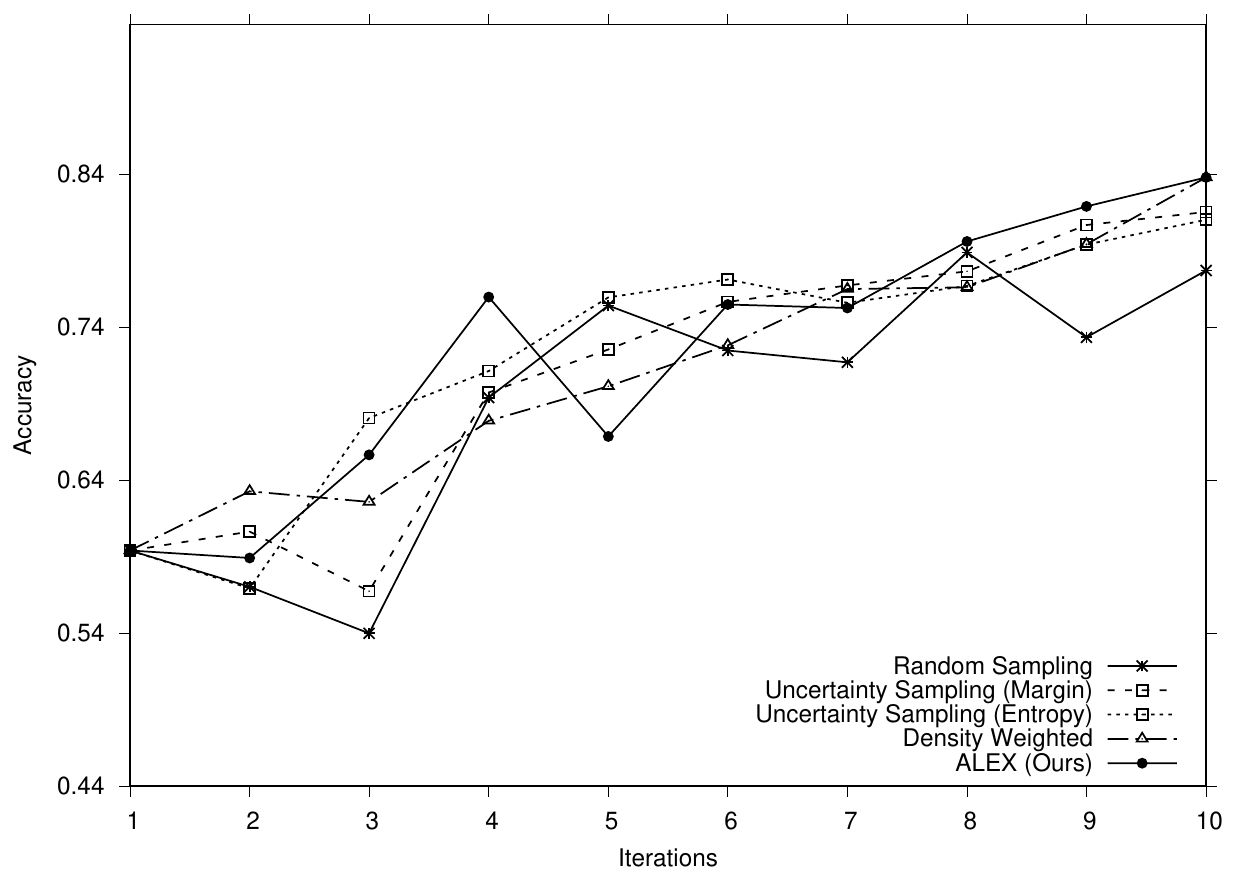}
    \caption{$q=5$ ($|S|=50$)}
\end{subfigure}
\begin{subfigure}[t]{.32\columnwidth}
    \centering
    \includegraphics[width=\columnwidth]{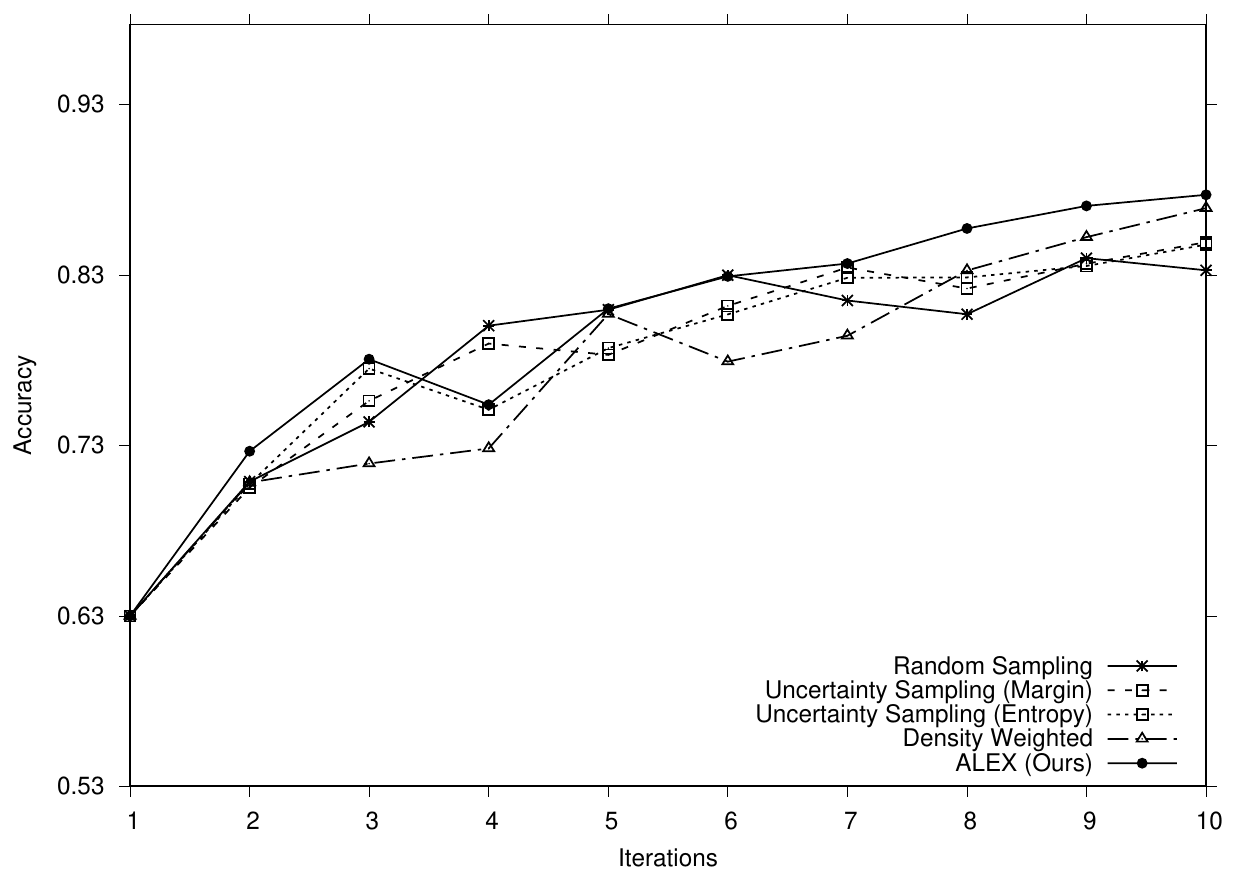}
    \caption{$q=10$ ($|S|=100$)}
\end{subfigure}
\caption{AL bootstrapping accuracy for MNIST with different values for initial seed sizes.}
\label{fig:mnist}
\end{figure}

\begin{figure}[t]
    \centering
\begin{subfigure}[t]{.32\columnwidth}
    \centering
    \includegraphics[width=\columnwidth]{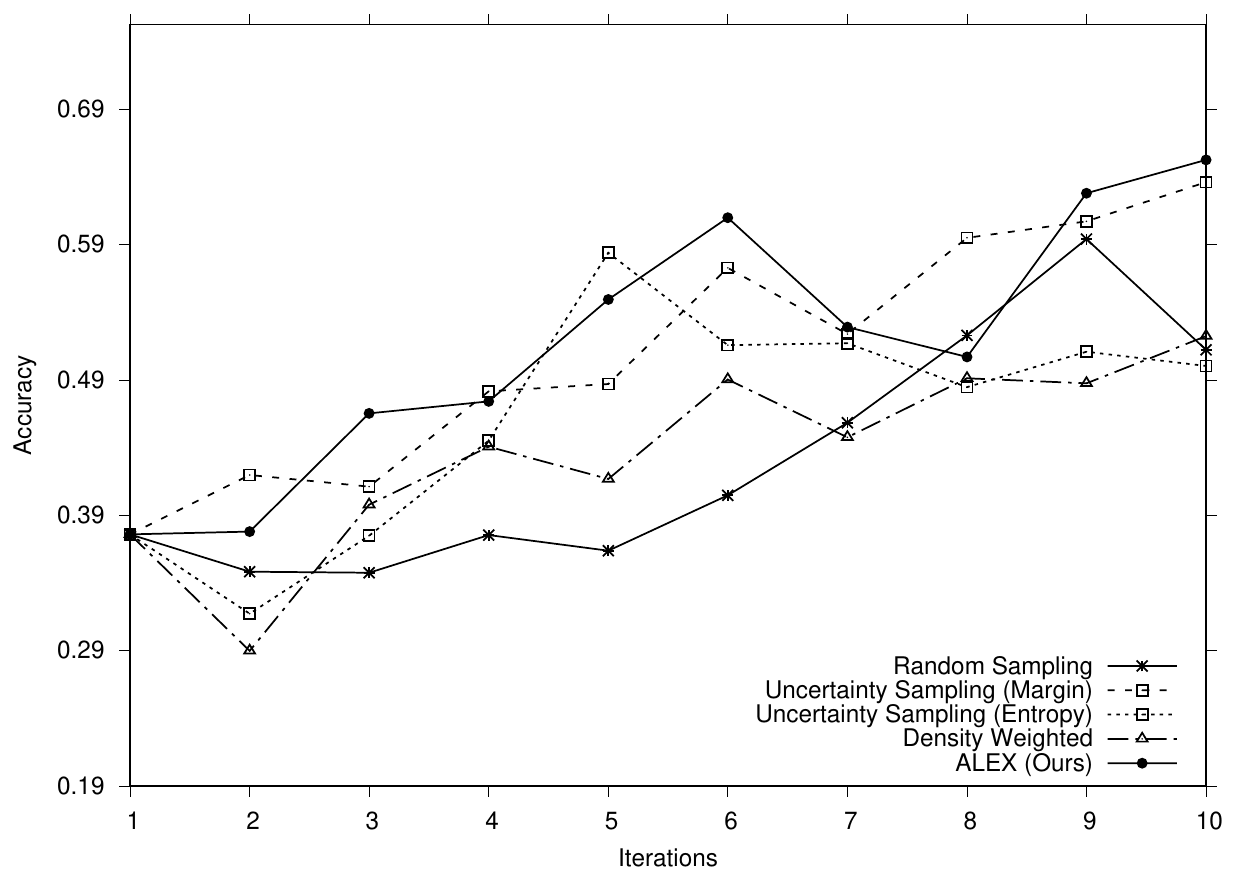}
    \caption{$q=1$ ($|S|=10$)}
\end{subfigure}
\begin{subfigure}[t]{.32\columnwidth}
    \centering
    \includegraphics[width=\columnwidth]{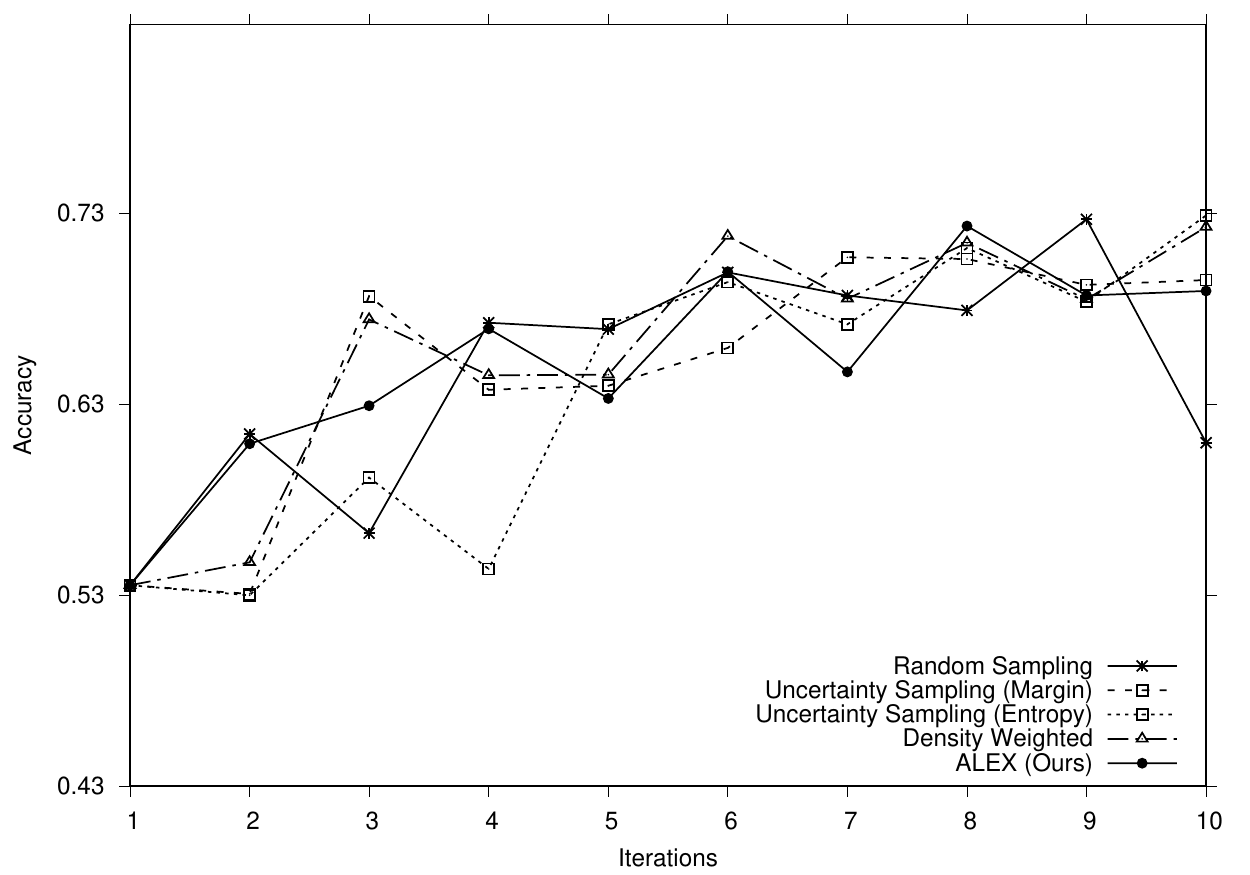}
    \caption{$q=5$ ($|S|=50$)}
\end{subfigure}
\begin{subfigure}[t]{.32\columnwidth}
    \centering
    \includegraphics[width=\columnwidth]{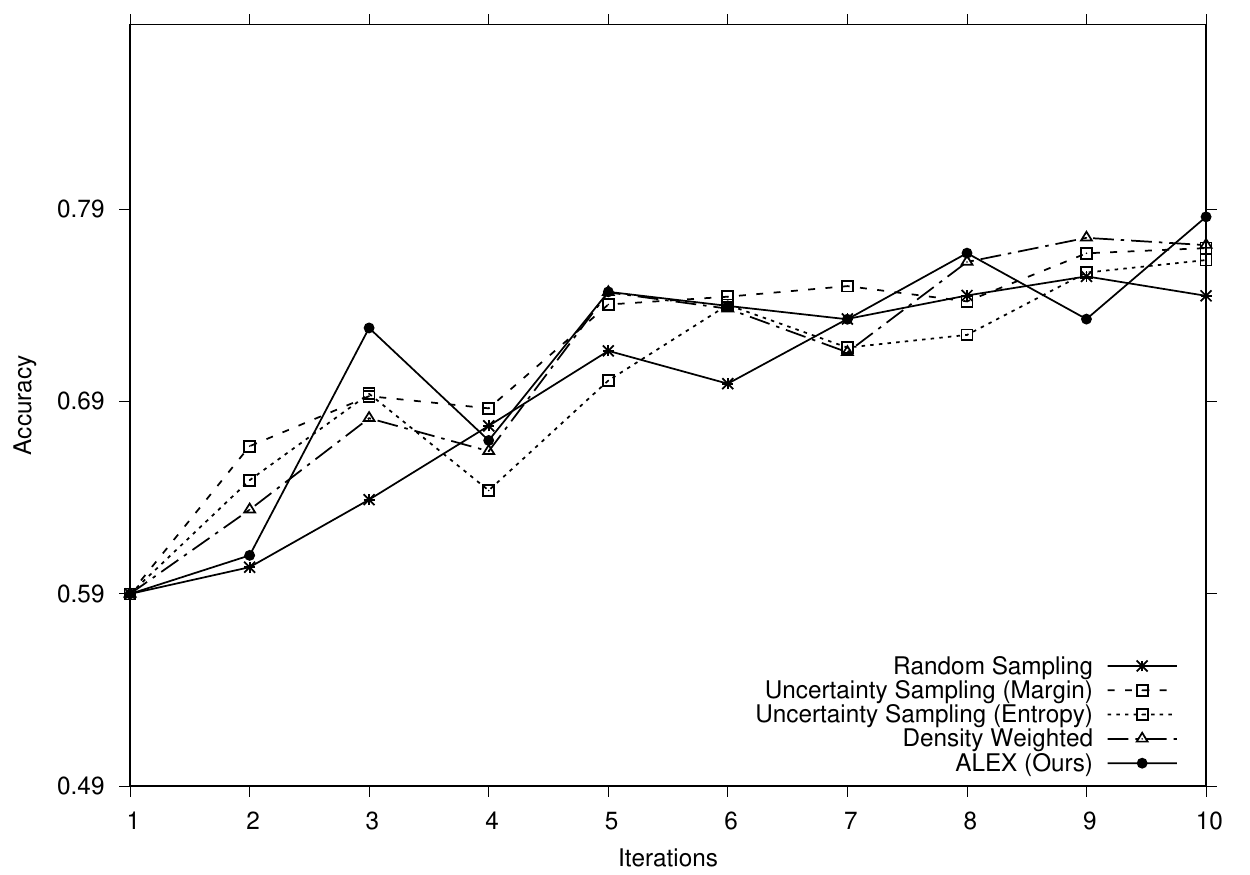}
    \caption{$q=10$ ($|S|=100$)}
\end{subfigure}
\caption{AL bootstrapping accuracy for FMNIST with different values for initial seed sizes.}
\label{fig:fmnist}
\end{figure}

Table \ref{tab:results} presents an overview of the results in terms of model effectiveness after $10$ bootstrapping steps with different AL methods with different number of instances per class to start with (the parameter $q$). Particularly interesting is to observe that even with a single labelled example per class, ALEX consistently outperforms the other baseline methods in terms of the test-set accuracy. This trend is mostly consistent with a higher number of initial seed points as well.   
To investigate the model effectiveness during the intermediate bootstrapping steps, we plot the accuracy (test-set) values for each AL iteration in Figures \ref{fig:mnist} and \ref{fig:fmnist}.

%
It can be observed that the curve corresponding to ALEX lies mostly above the other baseline AL methods (for both the datasets) which indicates that the hypothesis of annotating instances that are difficult to explain for the current classifier model is, in general, able to enhance its effectiveness during the next iteration. 


\para{Explanation Visualization}

\begin{figure}[t]
\centering
\begin{subfigure}[t]{.32\columnwidth}
\centering
\includegraphics[width=\columnwidth]{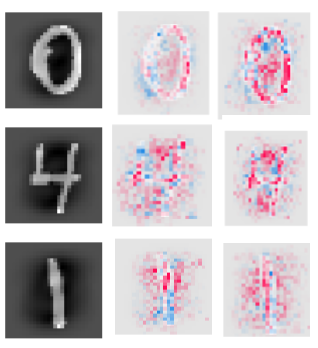}
\caption{MNIST  \label{fig:mnist_shap}}
\end{subfigure}
\quad
\begin{subfigure}[t]{.32\columnwidth}
\centering
\includegraphics[width=\columnwidth]{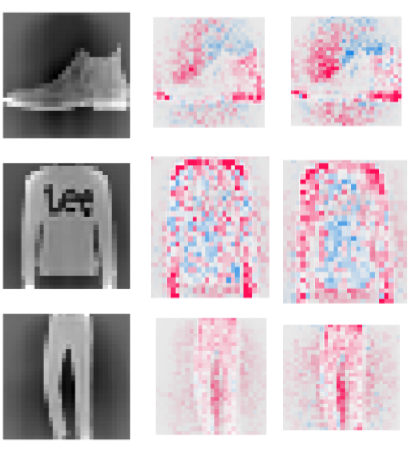}
\caption{FMNIST \label{fig:fmnist_shap}}
\end{subfigure}
\caption{AL model explanation of a sample gray-scale image (left column) for US with posteriors (middle column) and ALEX (right column) of each figure.
The warmth of a color (reddish hues) indicate higher weights of $\phi\cdot\vec{x}$ (Equation \ref{eq:instance_exp}).
}
\label{fig:shapplots}
\end{figure}

To demonstrate model interpretability in terms of the Shap method (combination of Shapley values for feature importance in combination with Equation \ref{eq:instance_exp}), we render the explanation vectors for sample images from the two datasets in Figure \ref{fig:shapplots}. From a visual perspective, the regions in an image with a reddish hue corresponds to parts of the images that the model leverages to make its predictions. With respect the MNIST classes, it can be seen that the reddish hues are distributed in regions that better correlate with human perception of identifying a hand-written digit. For example, to identify a `$0$', the model trained with US-P mainly focuses on the right part of the oval (reddish hues towards the right part of the image), whereas the classifier model trained with the ALEX AL method focuses on the whole oval instead. In the FMNIST dataset (Figure \ref{fig:fmnist_shap}) it can be seen that identification of an ankle-boot relies more on the shape of the ankle as well as on the flatness of the heel in ALEX, whereas for the baseline US-P, the decision is mainly based on the rear part of the shoe (which can in fact lead to misclassify an ankle boot as a sneaker).

\para{Concluding Remarks}
In this paper, we presented our work-in-progress towards developing not only effective but more interpretable classifier models with a small number of labeled data instances with the help of active learning based bootstrapping. Our experiments showed that training an explainer model (e.g. Shap \cite{SHAP}) to obtain the feature weights helps to formulate a new AL selection heuristic where the unlabeled instances that are chosen to be annotated are the ones that yield a different explanation, on an average, with respect to the already labeled ones.
With this set of encouraging results, in future we will conduct more experiments on other modalities of data (e.g. text and mixed modalities) and on other downstream tasks (e.g. sequence labeling) to see if our proposed AL heuristic can lead to similar observations.

{
\scriptsize
\bibliographystyle{ACM-Reference-Format}
\bibliography{alex}
}
\end{document}